%% file: egpaper_for_review.tex
\documentclass[10pt,twocolumn,letterpaper]{article}

\usepackage{iccv}
\usepackage{times}
\usepackage{epsfig}
\usepackage{graphicx}
\usepackage{amsmath}
\usepackage{amssymb}
\usepackage{booktabs}

\usepackage[pagebackref=true,breaklinks=true,letterpaper=true,colorlinks,bookmarks=false]{hyperref}

\usepackage[belowskip=-12pt,aboveskip=1pt]{caption}
\iccvfinalcopy 

\def\iccvPaperID{673} 

\newcommand{\zzhu}[1]{{#1}}

\begin{document}
	
	\title{Asymmetric Non-local Neural Networks for Semantic Segmentation}
	
	\author{Zhen~Zhu$^{1}$\thanks{~Equal contribution} , Mengde Xu$^{1}$\footnotemark[\value{footnote}] , Song Bai$^{2}$, Tengteng Huang$^1$, Xiang~Bai$^1$\thanks{~Corresponding author}\\
$^1${\em Huazhong University of Science and Technology}, $^2${\em University of Oxford}\\ 
{\tt \small \{zzhu, mdxu, huangtengtng, xbai\}@hust.edu.cn}, {\tt \small songbai.site@gmail.com}
	}
	\maketitle
	

	\begin{abstract}
		The non-local module works as a particularly useful technique for semantic segmentation while criticized for its prohibitive computation and GPU memory occupation. In this paper, we present \textbf{Asymmetric Non-local Neural Network} to semantic segmentation, which has two prominent components: Asymmetric Pyramid Non-local Block (APNB) and Asymmetric Fusion Non-local Block (AFNB). APNB leverages a pyramid sampling module into the non-local block to largely reduce the computation and memory consumption without sacrificing the performance. AFNB is adapted from APNB to fuse the features of different levels under a sufficient consideration of long range dependencies and thus considerably improves the performance. Extensive experiments on semantic segmentation benchmarks demonstrate the effectiveness and efficiency of our work. In particular, we report the state-of-the-art performance of 81.3 mIoU on the Cityscapes test set. For a $256 \times 128$ input, APNB is around 6 times faster than a non-local block on GPU while 28 times smaller in GPU running memory occupation. Code is available at: \url{https://github.com/MendelXu/ANN.git}.
		
	\end{abstract}
	
	\vspace{-4mm}
	
	\section{Introduction}
	
	Semantic segmentation is a long-standing challenging task in computer vision, aiming to predict pixel-wise semantic labels in an image accurately. This task is exceptionally important to tons of real-world applications, such as autonomous driving \cite{automated,autonomous}, medical diagnosing \cite{zhou2019prior,zhou2019semi}, \etc. 
	In recent years, the developments of deep neural networks encourage the emergence of a series of works~\cite{segnet,deeplabv1,refinenet,FCN,dilation,encnet,pspnet}. Shelhamer \emph{et al.} \cite{FCN} proposed the seminal work called Fully Convolutional Network (FCN), which discarded the fully connected layer to support input of arbitrary sizes. Since then, a lot of works \cite{deeplabv1,refinenet} were inspired to manipulate FCN techniques into deep neural networks. Nonetheless, the segmentation accuracy is still far from satisfactory.
	
	Some recent studies \cite{parsenet,nonlocal,pspnet} indicate that the performance could be improved if making sufficient use of long range dependencies. However, models that solely rely on convolutions exhibit limited ability in capturing these long range dependencies. A possible reason is the receptive field of a single convolutional layer is inadequate to cover correlated areas. Choosing a big kernel or composing a very deep network is able to enlarge the receptive field. However, such strategies require extensive computation and parameters, thus being very inefficient \cite{sagan}. Consequently, several works \cite{nonlocal,pspnet} resort to use global operations like non-local means \cite{nonlocalmean} and spatial pyramid pooling \cite{sppdetection,SpatialPyramidMatching}.

	\begin{figure}[tb]
		\centering
		\includegraphics[width=0.48\textwidth]{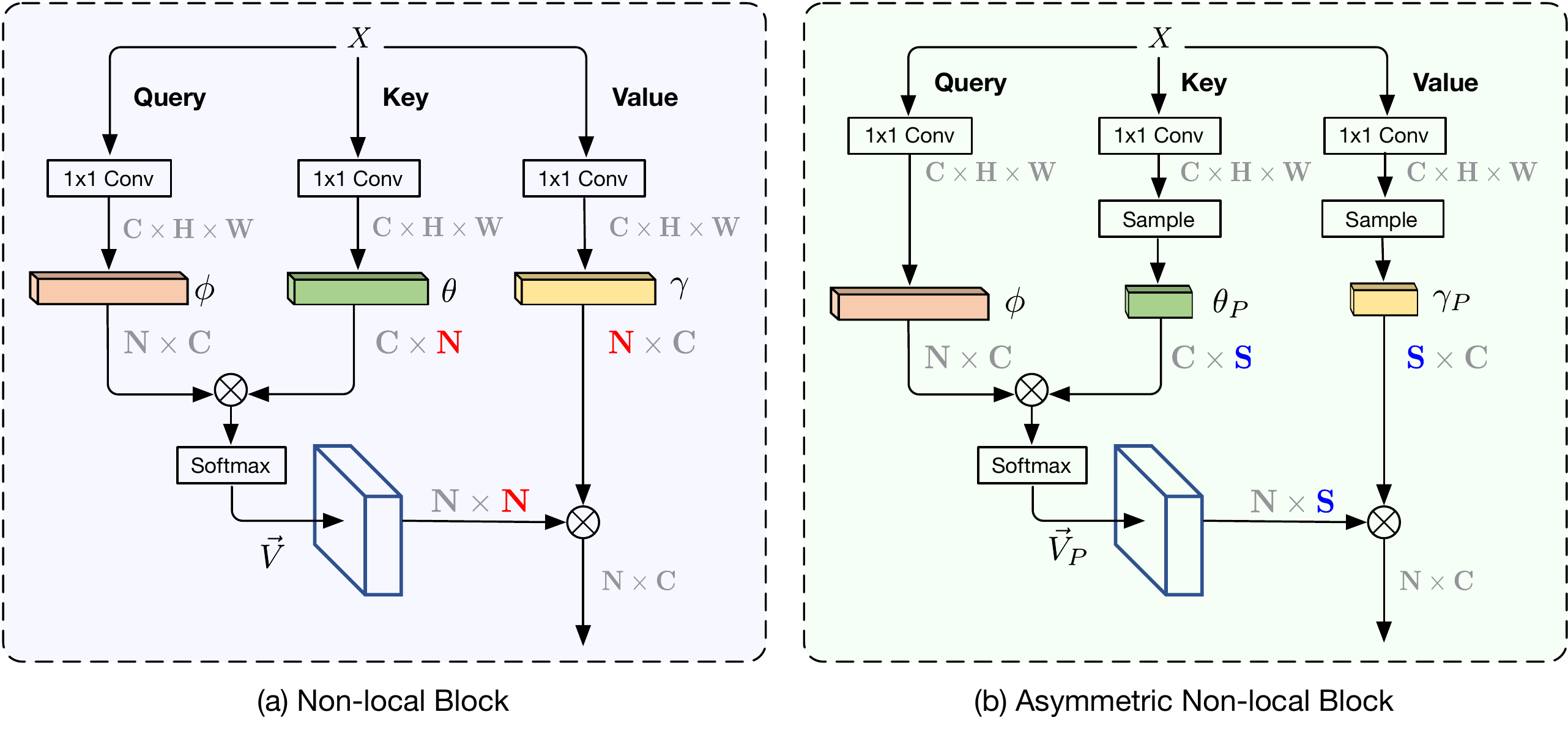}
		\caption{Architecture of a standard non-local block (a) and the asymmetric non-local block (b). $N=H\cdot W$ while $S \ll N$.}
		\label{fig:blocks}
	\end{figure}

	\begin{figure}[tb]
		\centering
		\includegraphics[width=0.48\textwidth]{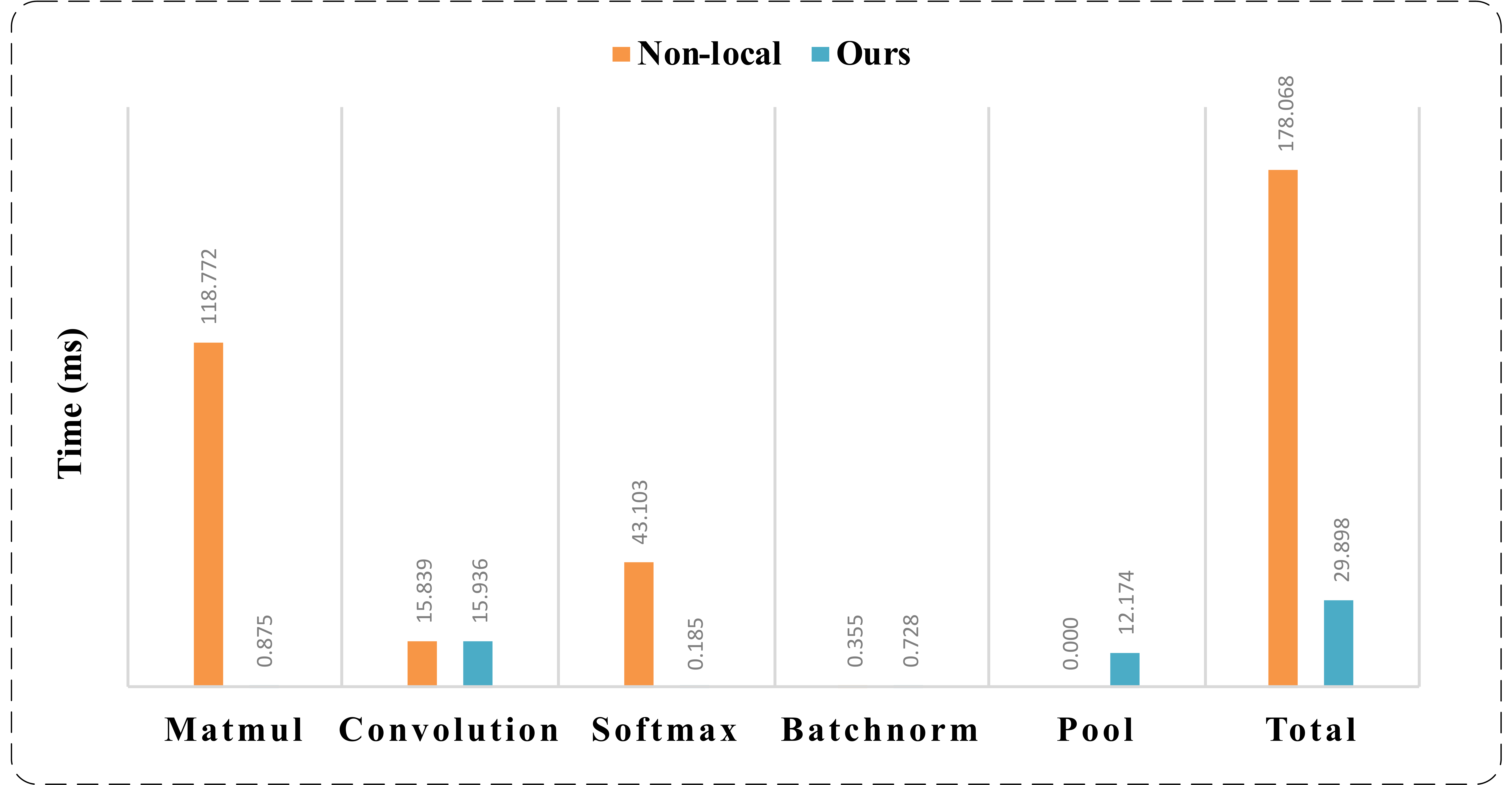}
		\caption{GPU time ( $\geq 1$ ms) comparison of different operations between a generic non-local block and our APNB. The last bin denotes the sum of all the time costs. The size of the inputs for these two blocks is $256 \times 128$.}
		\label{fig:time}
	\end{figure}

	In \cite{nonlocal}, Wang~\emph{et al.}~combined CNNs and traditional non-local means \cite{nonlocalmean} to compose a network module named non-local block in order to leverage features from all locations in an image. This module improves the performance of existing methods \cite{nonlocal}. However, the prohibitive computational cost and vast GPU memory occupation hinder its usage in many real applications. The architecture of a common non-local block \cite{nonlocal} is depicted in Fig.~\ref{fig:blocks}(a). The block first calculates the similarities of all locations between each other, requiring a matrix multiplication of computational complexity $\mathcal{O}(CH^{2}W^{2})$, given an input feature map with size $C\times H \times W$. Then it requires another matrix multiplication of computational complexity $\mathcal{O}(CH^{2}W^{2})$ to gather the influence of all locations to themselves. Concerning the high complexity brought by the matrix multiplications, we are interested in this work if there are efficient ways to solve this without sacrificing the performance.
	
	We notice that as long as the outputs of the \emph{key} branch and \emph{value} branch hold the same size, the output size of the non-local block remains unchanged. Considering this, if we could sample only a few representative points from \emph{key} branch and \emph{value} branch, it is possible that the time complexity is significantly decreased without sacrificing the performance. This motivation is demonstrated in Fig.~\ref{fig:blocks} when changing a large value $\textcolor{red}{N}$ in the \emph{key} branch and \emph{value} branch to a much smaller value $\textcolor{blue}{S}$ (From (a) to (b)). 
	
	
	
	In this paper, we propose a simple yet effective non-local module called \textbf{A}symmetric \textbf{P}yramid \textbf{N}on-local \textbf{B}lock (APNB) to decrease the computation and GPU memory consumption of the standard non-local module \cite{nonlocal} with applications to semantic segmentation. Motivated by the spatial pyramid pooling~\cite{sppdetection,SpatialPyramidMatching,pspnet} strategy, we propose to embed a pyramid sampling module into non-local blocks, which could largely reduce the computation overhead of matrix multiplications yet provide substantial semantic feature statistics. This spirit is also related to the sub-sampling tricks~\cite{nonlocal} (\eg,~max pooling). Our experiments suggest that APNB yields much better performance than those sub-sampling tricks with a decent decrease of computations.
	To better illustrate the boosted efficiency, we compare the GPU times of APNB and a standard non-local block in Fig.~\ref{fig:time}, averaging the running time of 10 different runs with the same configuration. Our APNB largely reduces the time cost on matrix multiplications, thus being nearly 6 times faster than a non-local block.
	
	Besides, we also adapt APNB to fuse the features of different stages of a deep network, which brings a considerable improvement over the baseline model. We call the adapted block as \textbf{A}symmetric \textbf{F}usion \textbf{N}on-local \textbf{B}lock (AFNB). AFNB calculates the correlations between every pixel of the low-level and high-level feature maps, yielding a fused feature with long range interactions. Our network is built based on a standard ResNet-FCN model by integrating APNB and AFNB together.
	
	
	The efficacy of the proposed network is evaluated on Cityscapes \cite{cityscapes}, ADE20K \cite{ade20k} and PASCAL Context \cite{pascal_context}, achieving the state-of-the-art performance 81.3\%, 45.24\% and 52.8\%, respectively. In terms of time and space efficiency, APNB is around 6 times faster than a non-local block on a GPU while 28 times smaller in GPU running memory occupation for a $256 \times 128$ input.
	
	
	
	\begin{figure*}[tb]
		\centering
		\includegraphics[width=0.92\textwidth]{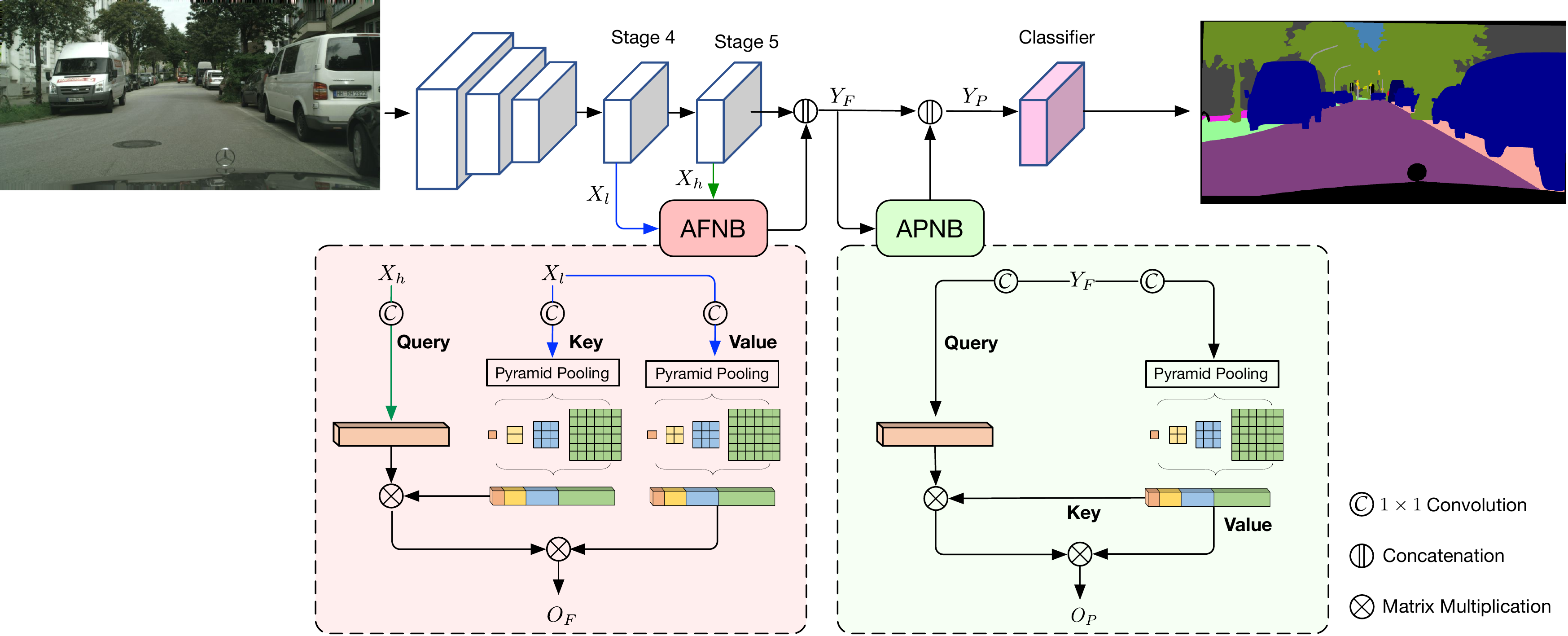}
		\caption{Overview of the proposed Asymmetric Non-local Neural Network. In our implementation, the \emph{key} branch and the \emph{value} branch in APNB share the same $1 \times 1$ convolution and sampling module, which decreases the number of parameters and computation without sacrificing the performance.}
		\label{fig:model}
	\end{figure*}

	\section{Related Work}
	In this section, we briefly review related works about semantic segmentation or scene parsing.  Recent advances focus on exploring the context information and can be roughly categorized into five directions:
	
	\vspace{1ex}\noindent \textbf{Encoder-Decoder.} A encoder generally reduces the spatial size of feature maps to enlarge the receptive field. Then the encoded codes are fed to the decoder, which is responsible for recovering the spatial size of the prediction maps. Long \emph{et al.} \cite{FCN} and Noh \emph{et al.} \cite{deconv} used deconvolutions to perform the decoding pass. Ronneberger \emph{et al.} \cite{unet} introduced skip-connections to bridge the encoding features to their corresponding decoding features, which could enrich the segmentation output with more details. Zhang \emph{et al.} \cite{encnet} introduced a context encoding module to predict semantic category importance and selectively strengthen or weaken class-specific feature maps.
	
	\vspace{1ex}\noindent \textbf{CRF.} As a frequently-used operation that could leverage context information in machine learning, Conditional Random Field \cite{crf} meets its new opportunity in combining with CNNs for semantic segmentation \cite{DeepGaussianCRFs,deeplabv1,deeplabv2,crfrnn,gcrf}. CRF-CNN \cite{crfrnn} adopted this strategy, making the deep network end-to-end trainable. Chandra \emph{et al.} \cite{DeepGaussianCRFs} and Vemulapalli \emph{et al.} \cite{gcrf} integrated Gaussian Conditional Random Fields into CNNs and achieved relatively good results.
	
	\vspace{1ex}\noindent \textbf{Different Convolutions. } Chen \emph{et al.} \cite{deeplabv1,deeplabv2} and Yu \emph{et al.} \cite{dilation} adapted generic convolutions to dilated ones, making the networks sensitive to global context semantics and thus improves the performance. Peng \emph{et al.} \cite{largekernelmatters} found large kernel convolutions help relieve the contradiction between classification and localization in segmentation. 
	
	\vspace{1ex}\noindent \textbf{Spatial Pyramid Pooling.} Inspired by the success of spatial pyramid pooling in object detection \cite{sppdetection}, Chen \emph{et al.} \cite{deeplabv2} replaced the pooling layers with dilated convolutions of different sampling weights and built an Atrous Spatial Pyramid Pooling layer (ASPP) to account for multiple scales explicitly. Chen \emph{et al.} \cite{deeplabv3+} further combined ASPP and the encoder-decoder architecture to leverage the advantages of both and boost the performance considerably.
	Drawing inspiration from \cite{SpatialPyramidMatching}, PSPNet \cite{pspnet} conducted spatial pyramid pooling after a specific layer to embed context features of different scales into the networks. Recently, Yang \emph{et al.} \cite{denseaspp} pointed out the ASPP layer has a restricted receptive field and adapted ASPP to a densely connected version, which helps to overcome such limitation. 
	
	\vspace{1ex}\noindent \textbf{Non-local Network.} Recently, researchers \cite{parsenet,nonlocal,pspnet} noticed that skillful leveraging the long range dependencies brings great benefits to semantic segmentation. Wang \emph{et al.} \cite{nonlocal} proposed a non-local block module combining non-local means with deep networks and showcased its efficacy for segmentation. 
	
	\vspace{1ex} Different from these works, our network uniquely incorporates pyramid sampling strategies with non-local blocks to capture the semantic statistics of different scales with only a minor budget of computation, while maintaining the excellent performance as the original non-local modules.
	

	\section{Asymmetric Non-local Neural Network}
	In this section, we firstly revisit the definition of non-local block~\cite{nonlocal} in Sec.~\ref{sec:revsit}, then detail the proposed Asymmetrical Pyramid Non-local Block ({\bf APNB}) and Asymmetrical Fusion Non-local Block ({\bf AFNB}) in Sec.~\ref{sec:APNB} and Sec.~\ref{sec:AFNB}, respectively. While APNB aims to decrease the computational overhead of non-local blocks, AFNB improves the learning capacity of non-local blocks thereby improving the segmentation performance. 
	
	
	\subsection{Revisiting Non-local Block} \label{sec:revsit}
	A typical non-local block~\cite{nonlocal} is shown in Fig.~\ref{fig:blocks}. Consider an input feature $X \in \mathcal{R}^{C \times H \times W}$, where $C$, $W$, and $H$ indicate the channel number, spatial width and height, respectively. Three $1\times1$ convolutions $W_{\phi}$, $W_{\theta}$, and $W_{\gamma}$ are used to transform $X$ to different embeddings $\phi \in {R}^{\hat{C} \times H \times W}$, $\theta \in \mathcal{R}^{\hat{C} \times H \times W}$ and $\gamma \in \mathcal{R}^{\hat{C} \times H \times W}$ as
	\begin{equation}
	\phi = W_{\phi}(\mathcal{X}),~~
	\theta = W_{\theta}(\mathcal{X}),~~ 
	\gamma = W_{\gamma}(\mathcal{X}),
	\end{equation}
	where $\hat{C}$ is the channel number of the new embeddings. 
	Next, the three embeddings are flattened to size $\hat{C} \times N$, where $N$ represents the total number of the spatial locations, that is, $N=H\cdot W$.
	Then, the similarity matrix $V \in \mathcal{R}^{N \times N}$ is calculated by a matrix multiplication as
	\begin{equation}
	V = \phi^\mathrm{T}\times\theta.
	\label{matmul1}
	\end{equation} 
	Afterward, a normalization is applied to $V$ to get a unified similarity matrix as
	\begin{equation}
	\vec{V} = f(V).
	\end{equation} 
	According to \cite{nonlocal}, the normalizing function $f$ can take the form from $\mathrm{softmax}$, $\mathrm{rescaling}$, and $\mathrm{none}$. We choose $\mathrm{softmax}$ here, which is equivalent to the self-attention mechanism and proved to work well in many tasks such as machine translation \cite{self-attention} and image generation \cite{sagan}. For every location in $\gamma$, the output of the attention layer is
	\begin{equation}
	O =\vec{V} \times \gamma^\mathrm{T},
	\label{matmul2}
	\end{equation} where $O  \in \mathcal{R}^{N \times \Hat{C}}$.
	By referring to the design of the non-local block, the final output is given by
	\begin{equation}
	Y = W_o(O^\mathrm{T}) + X \ \mathrm{or} \ Y = \mathrm{cat}(W_o(O^\mathrm{T}), X),
	\end{equation} 
	where $W_o$, also implemented by a $1\times1$ convolution, acts as a weighting parameter to adjust the importance of the non-local operation \emph{\wrt} the original input $X$ and moreover, recovers the channel dimension from $\hat{C}$ to $C$.
	
	\subsection{Asymmetric Pyramid Non-local Block} \label{sec:APNB}
	\label{apnb}
	The non-local network is potent to capture the long range dependencies that are crucial for semantic segmentation. However, the non-local operation is very time and memory consuming compared to normal operations in the deep neural network,~\eg,~convolutions and activation functions. 
	
	\vspace{1ex}\noindent\textbf{Motivation and Analysis.}~By inspecting the general computing flow of a non-local block, one could clearly find that Eq. \eqref{matmul1} and Eq. \eqref{matmul2} dominate the computation. The time complexities of the two matrix multiplications are both $\mathcal{O}(\hat{C}N^2)=\mathcal{O}(\hat{C}H^2W^2)$. In semantic segmentation, the output of the network usually has a large resolution to retain detailed semantic features \cite{deeplabv2,pspnet}. That means $N$ is large (for example in our training phase, $N=96 \times 96 = 9216$). Hence, the large matrix multiplication is the main cause of the inefficiency of a non-local block (see our statistic in Fig.~\ref{fig:time}).

	
	A more straightforward pipeline is given as
	\begin{equation}
		\underbrace{\mathcal{R}^{{N} \times \hat{C}} \times  \mathcal{R}^{\hat{C}\times \textcolor{red}{N}}}_{Eq. \eqref{matmul1}} \rightarrow \underbrace{\mathcal{R}^{{N} \times \textcolor{red}{N}} \times  \mathcal{R}^{\textcolor{red}{N} \times \hat{C}}}_{Eq. \eqref{matmul2}} \rightarrow \mathcal{R}^{N \times \hat{C}}.
	\end{equation}
	We hold a key yet intuitive observation that by changing $\textcolor{red}{N}$ to another number $\textcolor{blue}{S}$ ($\textcolor{blue}{S}\ll\textcolor{red}{N}$), the output size will remain the same, as
	\begin{equation}
		\mathcal{R}^{{N} \times \hat{C}} \times  \mathcal{R}^{\hat{C}\times \textcolor{blue}{S}} \rightarrow \mathcal{R}^{{N} \times \textcolor{blue}{S}} \times  \mathcal{R}^{\textcolor{blue}{S} \times \hat{C}} \rightarrow \mathcal{R}^{N \times \hat{C}}.
	\end{equation} 
	Returning to the design of the non-local block, changing $\textcolor{red}{N}$ to a small number $\textcolor{blue}{S}$ is equivalent to sampling several representative points from $\theta$ and $\gamma$ instead of feeding all the spatial points, as illustrated in Fig.~\ref{fig:blocks}. Consequently, the computational complexity could be considerably decreased.
	
	
	\begin{figure}[tb]
		\centering
		\includegraphics[width=0.45\textwidth]{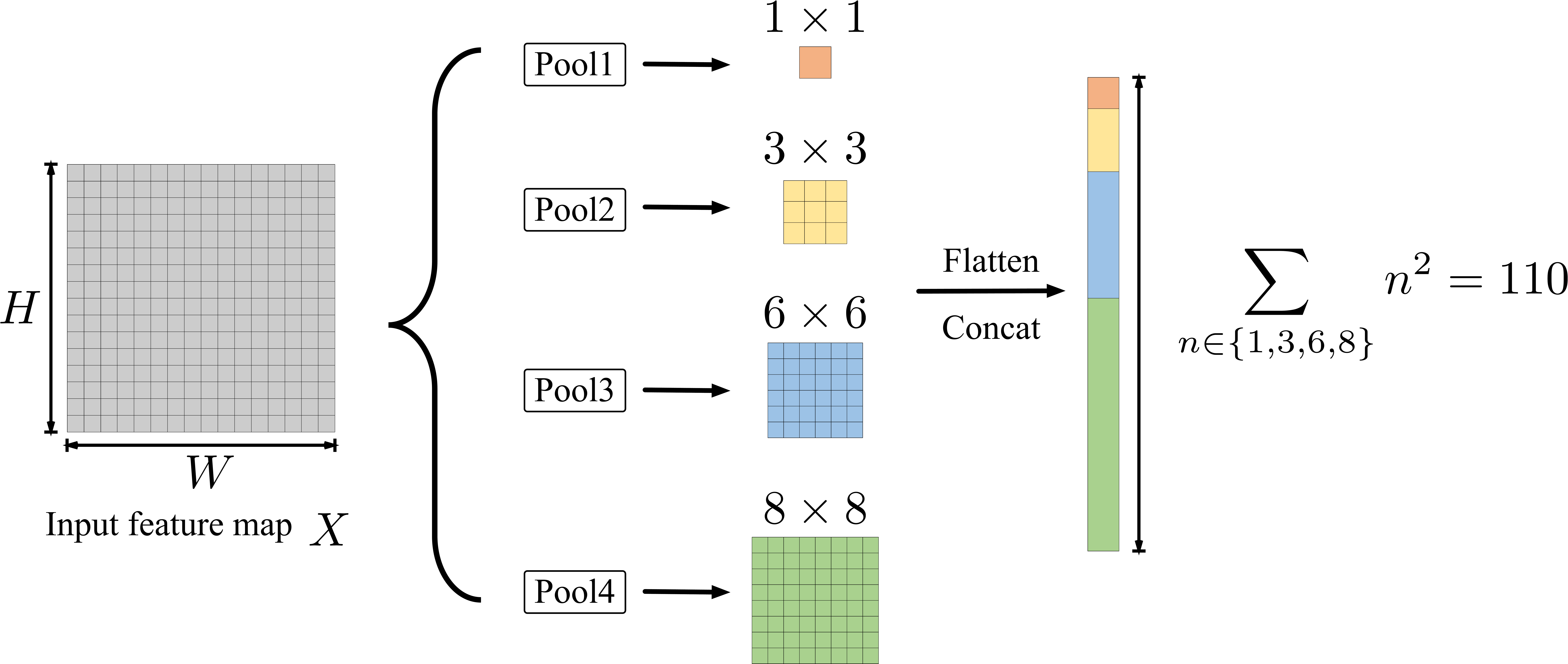}
		\caption{Demonstration of the pyramid max or average sampling process. }
		\label{fig:sampling_process}
	\end{figure}
	
	\vspace{1ex}\noindent\textbf{Solution.}~Based on the above observation, we propose to add sampling modules $\mathcal{P}_{\theta}$ and $\mathcal{P}_{\gamma}$ after $\theta$ and $\gamma$  to sample several sparse anchor points denoted as $\theta_{P} \in \mathcal{R}^{\hat{C} \times \textcolor{blue}{S}}$ and $\gamma_{P}\in \mathcal{R}^{\hat{C} \times \textcolor{blue}{S}}$, where $\textcolor{blue}{S}$ is the number of sampled anchor points. Mathematically, this is computed by
	\begin{equation}
		\theta_{P} = \mathcal{P}_{\theta}(\theta),~~\gamma_{P} = \mathcal{P}_{\gamma}(\gamma).
	\end{equation}
	The similarity matrix $V_{P}$ between $\phi$ and the anchor points $\theta_{P}$ is thus calculated by
	\begin{equation}
		V_{P} = \phi^\mathrm{T} \times \theta_{P}.
		\label{matmul_psp_1}
	\end{equation}
	Note that $V_{P}$ is an asymmetric matrix of size $N\times \textcolor{blue}{S}$.
	$V_{P}$ then goes through the same normalizing function as a standard non-local block, giving the unified similarity matrix $\vec{V}_P$. And the attention output is acquired by
	\begin{equation}
		O_P =\vec{V}_P \times {\gamma_{P}}^\mathrm{T},
		\label{matmul_psp_2}
	\end{equation} 
	where the output is in the same size as that of Eq.~\eqref{matmul2}. Following non-local blocks, the final output $Y_P$ is given as
	\begin{equation}
	Y_P = \mathrm{cat}(W_o({O_P}^\mathrm{T}), X).
	\end{equation} 
	
	The time complexity of such an asymmetric matrix multiplication is only $\mathcal{O}(\hat{C}NS)$, significantly lower than $\mathcal{O}(\hat{C}N^2)$ in a standard non-local block. It is ideal that $\textcolor{blue}{S}$ should be much smaller than $\textcolor{red}{N}$. However, it is hard to ensure that when $\textcolor{blue}{S}$ is small, the performance would not drop too much in the meantime. 

	
	As discovered by previous works \cite{SpatialPyramidMatching,pspnet}, global and multi-scale representations are useful for categorizing scene semantics. Such representations can be comprehensively carved by \emph{Spatial Pyramid Pooling} \cite{SpatialPyramidMatching}, which contains several pooling layers with different output sizes in parallel. In addition to this virtue, spatial pyramid pooling is also parameter-free and very efficient. Therefore, we embed pyramid pooling in the non-local block to enhance the global representations while reducing the computational overhead. 
	
	By doing so, we now arrive at the final formulation of Asymmetric Pyramid Non-local Block (APNB), as given in Fig.~\ref{fig:model}. As can be seen, our APNB derives from the design of a standard non-local block \cite{nonlocal}. A vital change is to add a spatial pyramid pooling module after $\theta$ and $\gamma$ respectively to sample representative anchors. 
	\zzhu{This sampling process is clearly depicted in Fig.~\ref{fig:sampling_process}, where several pooling layers are applied after $\theta$ or $\gamma$ and then the four pooling results are flattened and concatenated to serve as the input to the next layer.}
	We denote the spatial pyramid pooling modules as $\mathcal{P}_{\theta}^n$ and $\mathcal{P}_{\gamma}^n$, where the superscript $n$ means the width (or height) of the output size of the pooling layer (empirically, the width is equal to the height). In our model, we set $n\subseteq \{1, 3, 6, 8 \}$. Then the total number of the anchor points is
	
	\begin{equation}
		S=110=\sum_{n\in \{1, 3, 6, 8 \}}n^2.
	\end{equation} 
	As a consequence, the complexity of our asymmetric matrix multiplication is only 
	\begin{equation}
		T = \frac{S}{N}
		\label{eq:times}
	\end{equation} times of the complexity of the non-local matrix multiplication.
	When $H=128$ and $W=256$, the asymmetrical matrix multiplication saves us $\frac{256 \times 128}{110} \approx 298$ times of computation (see the results in Fig.~\ref{fig:time}).
	Moreover, the spatial pyramid pooling gives sufficient feature statistics about the global scene semantic cues to remedy the potential performance deterioration caused by the decreased computation. We will analyze this in our experiments.

	\subsection{Asymmetric Fusion Non-local Block} \label{sec:AFNB}
	Fusing features of different levels are helpful to semantic segmentation and object tracking as hinted in \cite{SpatialPyramidMatching,refinenet,FCN,deeplayeraggregation,ExFuse,similarityfusion}. Common fusing operations such as addition/concatenation, are conducted in a pixel-wise and local manner. We provide an alternative that leverages long range dependencies through a non-local block to fuse multi-level features, called Fusion Non-local Block.
	
	
	A standard non-local block only has one input source while FNB has two: a high-level feature map $X_h \in \mathcal{R}^{C_h \times N_h}$ and a low-level feature map $X_l \in \mathcal{R}^{C_l \times N_l}$. $N_h$ and $N_l$ are the numbers of spatial locations of $X_h$ and $X_l$, respectively. $C_h$ and $C_l$ are the channel numbers of $X_h$ and $X_l$, respectively. Likewise, $1 \times 1$ convolutions $W_{\phi}^h$, $W_{\theta}^l$ and $W_{\gamma}^l$ are used to transform $X_h$ and $X_l$ to embeddings ${\phi}_h \in \mathcal{R}^{\hat{C} \times N_h}$, ${\theta}_l \in \mathcal{R}^{\hat{C} \times N_l}$ and ${\gamma}_l \in \mathcal{R}^{\hat{C} \times N_l}$ as
	\begin{equation}
		{\phi}_h = W_{\phi}^h(\mathcal{X}_h), \ {\theta}_l = W_{\theta}^l(\mathcal{X}_l), \ {\gamma}_l = W_{\gamma}^l(\mathcal{X}_l).
	\end{equation}
	Then, the similarity matrix $V_F \in \mathcal{R}^{N_h \times N_l}$ between ${\phi}_h$ and ${\theta}_l$ is computed by a matrix multiplication
	\begin{equation}
		V_F  = {{\phi}_h}^{\mathrm{T}} \times {\theta}_l.
	\end{equation} 
	We also put a normalization upon $V_F$ resulting in a unified similarity matrix $\vec{V}_F \in \mathcal{R}^{N_h \times N_l}$. Afterward, we integrate $\vec{V}_F$ with $\gamma_l$ through a similar matrix multiplication as Eq. \eqref{matmul2} and Eq. \eqref{matmul_psp_2}, written as
	\begin{equation}
		O_F =\vec{V}_F \times {\gamma}_l^{\mathrm{T}}.
	\end{equation}
	The output $O_F \in \mathcal{R}^{N_h \times \hat{C}}$ reflects the bonus of $X_l$ to $X_h$, which are carefully selected from all locations in $X_l$. Likewise, $O_F$ is fed to a $1 \times 1$ convolution to recover the channel number to $C_h$. Finally, we have the output as
	\begin{equation}
		Y_F = \mathrm{cat}(W_o({O_F}^\mathrm{T}), X_h).
	\end{equation}
	
	Similar to the adaption of APNB \emph{\wrt} the generic non-local block, incorporating spatial pyramid pooling into FNB could derive an efficient Asymmetric Fusion Non-local Block (AFNB), as illustrated in Fig.~\ref{fig:model}. Inheriting from the advantages of APNB, AFNB is more efficient than FNB without sacrificing the performance.
	

	
	\subsection{Network Architecture}
	The overall architecture of our network is depicted in Fig.~\ref{fig:model}. We choose ResNet-101 \cite{resnet} as our backbone network following the choice of most previous works \cite{BiSeNet,pspnet,PSANet}. We remove the last two down-sampling operations and use the dilation convolutions instead to hold the feature maps from the last two stages\footnote{We refer to the stage with original feature map size $\frac{1}{16}$ as $\mathrm{Stage4}$ and
		size$\frac{1}{32}$ as $\mathrm{Stage5}$.}
	$\frac{1}{8}$ of the input image. Concretely, all the feature maps in the last three stages have the same spatial size. According to our experimental trials, we fuse the features of $\mathrm{Stage4}$ and $\mathrm{Stage5}$ using AFNB. The fused features are thereupon concatenated with the feature maps after $\mathrm{Stage5}$, avoiding situations that AFNB could not produce accurate strengthened features particularly when the training just begins and degrades the overall performance. Such features, full of rich long range cues from different feature levels, serve as the input to APNB, which then help to discover the correlations among pixels. As done for AFNB, the output of APNB is also concatenated with its input source. Note that in our implementation for APNB, $W_{\theta}$ and $W_{\gamma}$ share parameters in order to save parameters and computation, following the design of \cite{ocnet}. This design doesn't decrease the performance of APNB. Finally, a classifier is followed up to produce channel-wise semantic maps that later receive their supervisions from the ground truth maps. Note we also add another supervision to $\mathrm{Stage4}$ following the settings of \cite{pspnet}, as it is beneficial to improve the performance.
	
	\section{Experiments}
	To evaluate our method, we carry out detailed experiments on three semantic segmentation datasets: Cityscapes \cite{cityscapes}, ADE20K \cite{ade20k} and PASCAL Context \cite{pascal_context}. We have more competitive results on NYUD-V2 \cite{nyudv2} and COCO-Stuff-10K \cite{cocostuff} in the supplementary materials.

	
	\subsection{Datasets and Evaluation Metrics} 
	\noindent \textbf{Cityscapes} \cite{cityscapes} is particularly created for scene parsing, containing 5,000 high quality finely annotated images and 20,000 coarsely annotated images. All images in this dataset are shot on streets and of size $2048\times 1024$. The finely annotated images are divided into 2,975/500/1,525 splits for training, validation and testing, respectively. The dataset contains 30 classes annotations in total while only 19 classes are used for evaluation. 
	
	\vspace{1ex} \noindent \textbf{ADE20K} \cite{ade20k} is a large-scale dataset used in ImageNet Scene Parsing Challenge 2016, containing up to 150 classes. The dataset is divided into 20K/2K/3K images for training, validation, and testing, respectively.  Different from Cityscapes, both scenes and stuff are annotated in this dataset, adding more challenge for participated methods. 
	
	
	\vspace{1ex} \noindent \textbf{PASCAL Context} \cite{pascal_context} gives the segmentation labels of the whole image from PASCAL VOC 2010, containing 4,998 images for training and 5,105 images for validation. We use the 60 classes (59 object categories plus background) annotations for evaluation.

	\vspace{1ex}\noindent \textbf{Evaluation Metric.} We adopt Mean IoU (mean of class-wise intersection over union) as the evaluation metric for all the datasets. 
	
	\subsection{Implementation Details}
	\noindent \textbf{Training Objectives.}
	 Following \cite{pspnet}, our model has two supervisions: one after the final output of our model while another at the output layer of $\mathrm{Stage4}$. Therefore, our loss function is composed by two cross entropy losses as
	\begin{equation}
		\mathcal{L} = \mathcal{L}_{\mathrm{final}} + \lambda\mathcal{L}_{\mathrm{Stage4}}.
	\end{equation}
	For $\mathcal{L}_{\mathrm{final}}$, we perform online hard pixel mining, which excels at coping with difficult cases. $\lambda$ is set to 0.4.
	
	\vspace{1ex} \noindent \textbf{Training Settings.} Our code is based on an open source repository for semantic segmentation \cite{torchcv}  using PyTorch 1.0 framework \cite{pytorch}. The backbone network ResNet-101 is pretrained on the ImageNet \cite{Imagenet}. We use Stochastic Gradient Descent (SGD) to optimize our network, in which we set the initial learning rate to 0.01 for Cityscapes and PASCAL Context and 0.02 for ADE20K. During training, the learning rate is decayed according to the ``poly" leaning rate policy, where the learning rate is multiplied by $1 - (\frac{\mathrm{iter}}{\mathrm{max\_iter}})^{\mathrm{power}}$ with $\mathrm{power}=0.9$. For Cityscapes, we randomly crop out high-resolution patches $769 \times 769$ from the original images as the inputs for training \cite{deeplabv3,pspnet}. While for ADE20K and PASCAL Context, we set the crop size to $520 \times 520$ and $480 \times 480$, respectively \cite{encnet,pspnet}. 
	For all datasets, we apply random scaling in the range of [0.5, 2.0], random horizontal flip and random brightness as additional data augmentation methods.
	Batch size is 8 in Cityscapes experiments and 16 in the other datasets. We choose the cross-GPU synchronized batch normalization in \cite{encnet} or apex to synchronize the mean and standard-deviation of batch normalization layer across multiple GPUs. We also apply the auxiliary loss $\mathcal{L}_{\mathrm{Stage4}}$ and online hard example mining strategy in all the experiments as their effects for improving the performance are clearly discussed in previous works \cite{pspnet}. We train on the training set of Cityscapes, ADE20K and PASCAL Context for 60K, 150K, 28K iterations, respectively. All the experiments are conducted using $8\times$ Titan V GPUs.
	
	
	\vspace{1ex}\noindent \textbf{Inference Settings.}~For the comparisons with state-of-the-art methods, we apply multi-scale whole image and left-right flip testing for ADE20K and PASCAL Context while multi-scale sliding crop and left-right flip testing for the Cityscapes testing set. For quick ablation studies, we only employ single scale testing on the validation set of Cityscapes by feeding the whole original images.

	\subsection{Comparisons with Other Methods}
	
	\subsubsection{Efficiency Comparison with Non-local Block}
	As discussed in Sec.~\ref{sec:APNB}, APNB is much more efficient than a standard non-local block. We hereby give a quantitative efficiency comparison between our APNB and a generic non-local block in the following aspects: GFLOPs, GPU memory (\emph{MB}) and GPU computation time (\emph{ms}). In our network, non-local block/APNB receives a $96 \times 96$ (1/8 of the $769 \times 769$ input image patch) feature map during training while $256 \times 128$ (1/8 of the $2048 \times 1024$ input image) during single scale testing for Cityscapes dataset. Hence, we give relevant statistics of the two sizes.
	The testing environment is identical for these two blocks, that is, a Titan Xp GPU under CUDA 9.0 without other ongoing programs. Note our APNB has four extra adaptive average pooling layers to count as opposed to the non-local block while other parts are entirely identical. The comparison results are given in Tab.~\ref{tab:efficiency_comparison}. Our APNB is superior to the non-local block in all aspects. Enlarging the input size will give a further edge to our APNB because in Eq. \eqref{eq:times}, $N$ grows quadratically while $S$ remains unchanged.
	
	\zzhu{Besides the comparison of the single block efficiency, we also provide the whole network efficiency comparison with the two most advanced methods, PSANet \cite{PSANet} and DenseASPP \cite{denseaspp}, in terms of inference time (\emph{s}), GPU occupation with batch size set to 1 (\emph{MB}) and the number of parameters (\emph{Million}). According to Tab.~\ref{tab:efficiency_comparison}, though our inference time and parameter number are larger than DenseASPP \cite{denseaspp}, the GPU memory occupation is obviously smaller. We attribute this to the different backbone networks: ResNet comparatively contains more parameters and layers while DenseNet is more GPU memory demanding. When comparing with the previous advanced method PSANet \cite{PSANet}, which shares the same backbone network with us, our model is more advantageous in all aspects. This verifies our network is superior because of the effectiveness of APNB and AFNB rather than just having more parameters than previous works.}
	
	
	\begin{table}[tb]
		\centering
		\scriptsize
		\begin{tabular}{l|c|l|l|l}
			\toprule[2pt]
			\hline
			Method          & Input size & GFLOPs & GPU memory & GPU time  \\ \hline
			NB        & $96\times 96$   & 58.0 &  609  & 19.5  \\
			APNB  &  $96\times 96$ & 15.5 ($\downarrow$ 42.5) & 150 ($\downarrow$ 459) & 12.4 ($\downarrow$ 7.1)  \\ \hline
			NB	& $256\times 128$ & 601.4  & 7797  & 179.4    \\
			APNB &  $256\times 128$  & 43.5  ($\downarrow$ 557.9) & 277 ($\downarrow$ 7520)   & 30.8 ($\downarrow$ 148.6)   \\ \hline
			\bottomrule[2pt]
		\end{tabular}
		\caption{Computation and memory statistics comparisons between non-local block and our APNB. The channel numbers of the input feature maps $X$ is $C=2048$ and of the embeddings $\phi, \phi_P$ \etc is $\hat{C}=256$, respectively. Batch size is 1. The lower values, the better. \label{tab:efficiency_comparison}}
	\end{table}
	
\begin{table}[h]
	\centering
	\resizebox{0.48\textwidth}{!}{
	\begin{tabular}{l|l|l|l|l}
		\toprule[2pt]
		\hline
		Method          & Backbone & Inf. time (s) & Mem. (MB) & \# Param (M)  \\ \hline
		DenseASPP \cite{denseaspp}		& DenseNet-161  & 0.568 &  7973  & 35.63  \\
		PSANet \cite{PSANet}		      &  ResNet-101 & 0.672 & 5233  & 102.66  \\
		Ours			        & ResNet-101 & 0.611  & 3375  & 63.17     \\ \hline
		\bottomrule[2pt]
	\end{tabular}}
	\caption{Time, parameter and GPU memory comparisons based on the whole networks. \emph{Inf. time, Mem., \# Param} mean inference time, GPU memory occupation and number of parameters, respectively. Results are averaged from feeding ten 2048 $\times$ 1024 images. \label{tab:efficiency_comparison}}
\end{table}
	
	\subsubsection{Performance Comparisons} 
\begin{figure*}[tb]
		\centering
		\includegraphics[width=0.95\textwidth]{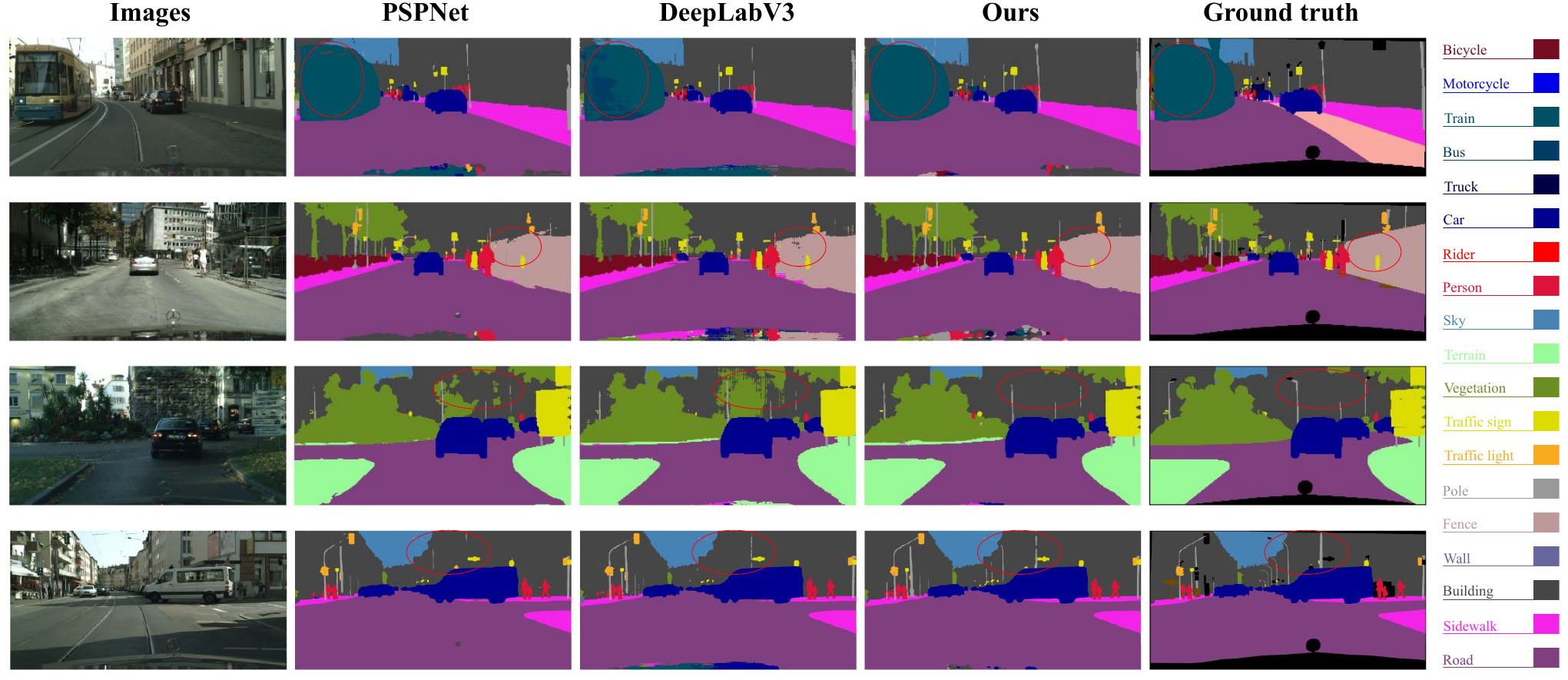}
		\caption{Qualitative comparisons with DeepLab-V3 \cite{deeplabv3} and PSPNet \cite{pspnet}. The \textcolor{red}{red} circles mark where our model is particularly superior to other methods.}
		\label{fig:comparison}
	\end{figure*}

	\noindent \textbf{Cityscapes.}~To compare the performance on the test set of Cityscapes with other methods, we directly train our asymmetric non-local neural network for 120K iterations with only the finely annotated data, including the training and validation sets. 
	As shown in Tab.~\ref{tab:cityscapes_test_comparison}, our method outperforms the previous state-of-the-art methods, attaining the performance of 81.3\%. \zzhu{We give several typical qualitative comparisons with other methods in Fig.~\ref{fig:comparison}. DeepLab-V3 \cite{deeplabv3} and PSPNet \cite{pspnet} are somewhat troubled with local inconsistency on large objects like truck (first row), fence (second row) and building (third row) \etc while our method isn't. Besides, our method performs better for very slim objects like the pole (fourth row) as well. }
	
	\begin{table}[tb]
		\centering
		\begin{tabular}{l|c|c|c}
			\toprule[2pt]
			\hline
			Method          & Backbone & Val & mIoU (\%)  \\ \hline
			DeepLab-V2 \cite{deeplabv2}       & ResNet-101   &       & 70.4    \\
			RefineNet \cite{refinenet}              & ResNet-101    &   \checkmark   & 73.6    \\
			GCN \cite{largekernelmatters}		& ResNet-101   &   \checkmark   & 76.9 \\
			DUC \cite{duc}       & ResNet-101    &    \checkmark    & 77.6   \\
			SAC \cite{sac}       & ResNet-101    &    \checkmark    & 78.1   \\
			ResNet-38 \cite{widerresnet}       & WiderResNet-38    &      & 78.4   \\
			PSPNet \cite{pspnet}       & ResNet-101    &     & 78.4   \\
			BiSeNet \cite{BiSeNet}       & ResNet-101    &    \checkmark    & 78.9   \\
			AAF \cite{aaf}       & ResNet-101    &    \checkmark    & 79.1   \\
			DFN \cite{dfn}       & ResNet-101    &    \checkmark    & 79.3   \\
			PSANet \cite{PSANet}       & ResNet-101    &    \checkmark    & 80.1   \\
			DenseASPP \cite{denseaspp}       & DenseNet-161    &    \checkmark    & 80.6   \\ \hline
			Ours       & ResNet-101    &   \checkmark    & \textbf{81.3}    \\ \hline
			\bottomrule[2pt]
		\end{tabular}
		\caption{Comparisons on the test set of Cityscapes with the state-of-the-art methods. Note that the \emph{Val} column indicates whether including the finely annotated validation set data of Cityscapes for training. \label{tab:cityscapes_test_comparison}}
	\end{table}
	
	\vspace{1ex} \noindent \textbf{ADE20K.}~As is known, ADE20K is challenging due to its various image sizes, lots of semantic categories and the gap between its training and validation set. Even under such circumstance, our method achieves better results than EncNet \cite{encnet}. It is noteworthy that our result is better than PSPNet \cite{pspnet} even when it uses a deeper backbone ResNet-269.
	
	\begin{table}[tb]
		\centering
		\begin{tabular}{l|c|c}
			\toprule[2pt]
			\hline
			Method          & Backbone  & mIoU (\%)  \\ \hline
			RefineNet \cite{refinenet}    & ResNet-152   & 40.70    \\
			UperNet \cite{upernet}    & ResNet-101   & 42.65    \\
			DSSPN \cite{dsspn}       & ResNet-101     &43.68   \\
			PSANet \cite{PSANet}    & ResNet-101     & 43.77   \\
			SAC \cite{sac}       & ResNet-101     & 44.30   \\
			EncNet \cite{encnet}       & ResNet-101     & 44.65   \\ 
			PSPNet \cite{pspnet}       & ResNet-101     & 43.29   \\
			PSPNet \cite{pspnet}       &ResNet-269     & 44.94 \\ \hline
			Ours       & ResNet-101   & \textbf{45.24}    \\ \hline
			\bottomrule[2pt]
		\end{tabular}
		\caption{Comparisons on the validation set of ADE20K with the state-of-the-art methods. \label{tab:ade20k_val_comparison}}
	\end{table}
	
	\vspace{1ex} \noindent \textbf{PASCAL Context.}~We report the comparison with state-of-the-art methods in Tab.~\ref{tab:pascalcontext_test_comparison}. It can be seen that our model achieves the state-of-the-art performance of 52.8\%. This result firmly suggests the superiority of our method. 
	
	
	\begin{table}[tb]
		\centering
		\begin{tabular}{l|c|c}
			\toprule[2pt]
			\hline
			Method          & Backbone  & mIoU (\%)  \\ \hline
			FCN-8s \cite{FCN}    &  --   & 37.8    \\
			Piecewise \cite{piecewise} & --  & 43.3 \\
			DeepLab-V2 \cite{deeplabv2}    & ResNet-101   & 45.7    \\
			RefineNet \cite{refinenet}       & ResNet-152     & 47.3   \\
			PSPNet \cite{pspnet}  & ResNet-101 & 47.8 \\
			CCL \cite{ccl}    & ResNet-101     & 51.6   \\
			EncNet \cite{encnet} & ResNet-101 & 51.7 \\ \hline
			Ours       & ResNet-101   & \textbf{52.8}    \\ \hline
			\bottomrule[2pt]
		\end{tabular}
		\caption{Comparisons on the validation set of PASCAL Context with the state-of-the-art methods. \label{tab:pascalcontext_test_comparison}}
	\end{table}
	

	\subsection{Ablation Study}
	
	\begin{table}[tb]
		\centering
		\begin{tabular}{l|c}
			\toprule[2pt]
			\hline
			Method         & mIoU (\%)  \\ \hline
			Baseline         & 75.8    \\
			+ NB      & 78.4    \\
			+ APNB		    & 78.6 \\ \hline
			+ Common fusion     & 76.5   \\
			+ FNB      & 77.3   \\
			+ AFNB      & 77.1   \\ \hline
			+ Common fusion + NB & 79.0 \\
			+ FNB + NB      & 79.7   \\
			+ AFNB + APNB (Full)    & 79.9   \\ \hline
			\bottomrule[2pt]
		\end{tabular}
		\caption{Ablation study on the validation set of Cityscapes about APNB and AFNB. ``+ $Module$'' means add ``$Module$'' to the Baseline model. \label{tab:cityscapes_val_ablation}}
	\end{table}

	In this section, we give extensive experiments to verify the efficacy of our main method. We also give several design choices and show their influences on the results. All the following experiments adopt ResNet-101 as the backbone, trained on the fine-annotated training set of Cityscapes for 60K iterations. 
	
	\vspace{1ex}\noindent \textbf{Efficacy of the APNB and AFNB.} Our network has two prominent components: APNB and AFNB. The following will evaluate the efficacy of each and the integration of both. The \textbf{Baseline} network is basically a FCN-like ResNet-101 network with a deep supervision branch. By adding a non-local block (\textbf{+ NB}) before the classifier to the \emph{Baseline} model, the performance is improved by 2.6\% (75.8\% $\rightarrow$ 78.4\%), as shown in Tab.~\ref{tab:cityscapes_val_ablation}. By replacing the normal non-local block with our APNB (\textbf{+ APNB}), the performance is slightly better (78.4\% $\rightarrow$ 78.6\%). 
	
	When adding a common fusion module (\textbf{+ Common Fusion}) from $\mathrm{Stage4}$ to $\mathrm{Stage5}$: $\mathrm{Stage5} + \mathrm{ReLU(BatchNorm(Conv(Stage4)))}$ to Baseline model, we also achieve a good improvement compared to the Baseline (75.8\% $\rightarrow$ 76.5\%). This phenomenon verifies the usefulness of the strategy that fuses the features from the last two stages. Replacing the common fusion module with our proposed Fusion Non-local Block (\textbf{+ FNB}), the performance is further boosted at 0.8\% (76.5\% $\rightarrow$ 77.3\%). Likewise, changing FNB to AFNB (\textbf{+ AFNB}) reduces the computation considerably at the cost of a minor performance decrease (77.3\% $\rightarrow$ 77.1\%). 
	
	To study whether the fusion strategy could further boost the highly competitive \emph{+ NB} model, we add \emph{Common fusion} to \emph{+ NB} model (\textbf{+ Common fusion + NB}) and achieve 0.6\% performance improvement (78.4\% $\rightarrow$ 79.0\%).
	Using both the fusion non-local block and typical non-local block (\textbf{+ FNB + NB}) can improve the performance of 79.7\%. 
	Using the combination of APNB and AFNB, namely our asymmetric non-local neural network (\textbf{+ AFNB + APNB (Full)} in Fig.~\ref{fig:model}), achieves the best performance of 79.9\%, demonstrating the efficacy of APNB and AFNB.

	\vspace{1ex}\noindent \textbf{Selection of Sampling Methods.} As discussed in Sec.~\ref{apnb}, the selection of the sampling module has a great impact on the performance of APNB. Normal sampling strategies include: \emph{max}, \emph{average} and \emph{random}. When integrated into spatial pyramid pooling, there goes another three strategies: \emph{pyramid max}, \emph{pyramid average} and \emph{pyramid random}. 
	We thereupon conduct several experiments to study their effects by combining them with APNB. As shown in Tab.~\ref{tab:pooling_ablation}, average sampling performs better than max and random sampling, which conforms to the conclusion drawn in \cite{pspnet}. We reckon it is because the resulted sampling points are more informative by receiving the provided information of all the input locations inside the average sampling kernel, when compared to the other two. This explanation could also be transferred to pyramid settings. Comparing average sampling and pyramid sampling under the same number of anchor points (third row \vs the last row), we can surely find pyramid pooling is a very key factor that contributes to the significant performance boost.
	
	\vspace{1ex}\noindent \textbf{Influence of the Anchor Points Numbers.} In our case, the output sizes of the pyramid pooling layers determine the total number of anchor points, which influence the efficacy of APNB. To investigate the influence, we perform the following experiments by altering the pyramid average pooling output sizes: $(1, 2, 3, 6)$, $(1, 3, 6, 8)$ and $(1, 4, 8, 12)$. As shown in Tab.~\ref{tab:pooling_ablation}, it is clear that more anchor points improve the performance with the cost of increasing computation. Considering this trade-off between efficacy and efficiency, we opt to choose $(1,3,6,8)$ as our default setting. 
	
	\begin{table}[tb]
		\centering
		\begin{tabular}{l|c|c|c}
			\toprule[2pt]
			\hline
			Sampling method  & $n$ & $\textcolor{blue}{S}$ & mIoU (\%)  \\ \hline
			random                   & 15      &  225 &  78.2   \\
			max                         & 15      &  225 &  78.1   \\
			average                   & 15      &  225 &  78.4  \\  \hline
			pyramid random   & 1,2,3,6  & 50  & 78.8   \\
			pyramid max   & 1,2,3,6  &  50 & 79.1   \\
			pyramid average    & 1,2,3,6 &  50  & 79.3   \\
			pyramid average    & 1,3,6,8  &  110 & 79.9   \\
			pyramid average    & 1,4,8,12 &  225  & 80.1   \\ \hline
			\bottomrule[2pt]
		\end{tabular}
		\caption{Ablation study on the validation set of Cityscapes in terms of sampling methods and anchor point number. ``$n$'' column represents the output width/height of a pooling layer. Note when implementing random and pyramid random, we use the $\mathrm{numpy.random.choice}$ function to randomly sample $n^2$ anchor points from all possible locations.
		``$S$'' column means the total number of the anchor points. \label{tab:pooling_ablation}}
	\end{table}
	
	\section{Conclusion}
	In this paper, we propose an asymmetric non-local neural network for semantic segmentation. The core contribution of asymmetric non-local neural network is the asymmetric pyramid non-local block, which can dramatically improve the efficiency and decrease the memory consumption of non-local neural blocks without sacrificing the performance. Besides, we also propose asymmetric fusion non-local block to fuse features of different levels. The asymmetric fusion non-local block can explore the long range spatial relevance among features of different levels, which demonstrates a considerable performance improvement over a strong baseline. Comprehensive experimental results on the Cityscapes, ADE20K and PASCAL Context datasets show that our work achieves the new state-of-the-art performance. In the future, we will apply asymmetric non-local neural networks to other vision tasks.
	
	\section*{Acknowledgement}
	This work was supported by NSFC 61573160, to Dr. Xiang Bai by the National Program for Support of Top-notch Young Professionals and the Program for HUST Academic Frontier Youth Team 2017QYTD08. We sincerely thank Huawei EI Cloud for their generous grant of GPU use for our paper. We genuinely thank Ansheng You for his kind help and suggestions throughout the project.

	{\small
		\bibliographystyle{ieee_fullname}
		\bibliography{egbib}
	}
	
	\clearpage
	\input{appendix.tex}
	
\end{document}

%% file: appendix.tex
\appendix
	\section{Quantitative comparisons on COCO-Stuff-10K and NYUD-V2}
	Following the training and evaluation protocols of DeepLab-V2 \cite{deeplabv2} and RefineNet \cite{refinenet}, our method achieves competitive results on the two datasets using single scale whole image testing, as shown in Tab.~\ref{tab:coco_and_nyud}. As NYUD-V2 \cite{nyudv2} is a small benchmark and COCO-Stuff-10K is quite challenging and large, these results further verify the effectiveness of our method on both small and large benchmarks.
	
	\begin{table}[h]
	\centering
    \resizebox{0.48\textwidth}{!}{
	\begin{tabular}{l|l|l||l|l|l}
		\hline
		Method       & Backbone   & mIoU (\%) & Method  & Backbone & mIoU (\%)  \\ \hline
		RefineNet \cite{refinenet}	 & ResNet-101	& 33.6  & Piecewise \cite{piecewise}  & VGG16  &  40.6  \\
		CCL \cite{ccl} & ResNet-101	& 35.7  & RefineNet \cite{refinenet} & ResNet-101  &  43.6    \\ \hline
		Ours			 &   ResNet-101    &  37.2 & Ours  &  ResNet-101 & 44.4   \\ \hline
	\end{tabular}}
	\caption{Comparisons on COCO-Stuff-10K (Left) and NYUD-V2 (Right) datasets. Results of the competing methods are taken from their papers.  \label{tab:coco_and_nyud}}
\end{table}
	
	\section{More ablation results}
	
\begin{figure*}[tb]
	\centering
	\includegraphics[width=\textwidth]{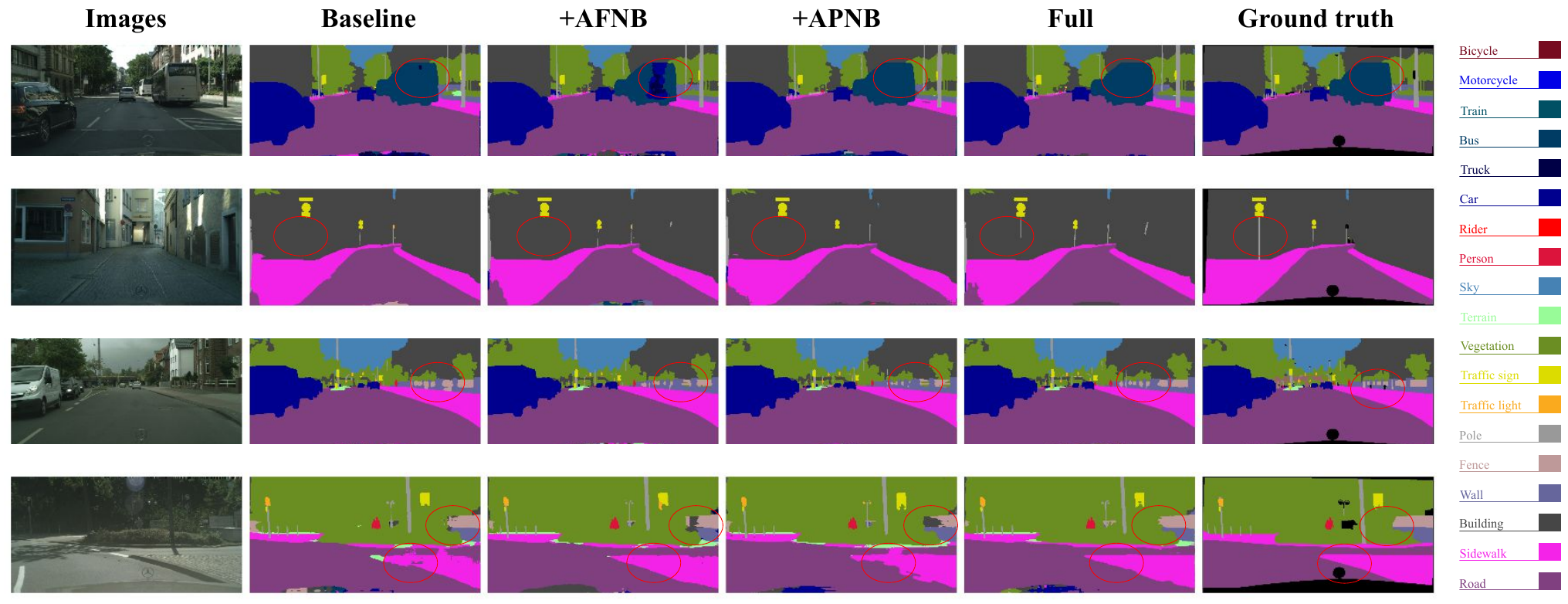}
	\caption{Qualitative comparisons among Full model and other variants of our model. The \textcolor{red}{red} circles indicate where Full model is superior to other model variants. \label{fig:self}}
    \end{figure*}

	\vspace{1ex}\noindent \textbf{Qualitative comparisons.} 
	We also give the qualitative comparisons of our Full (+ AFNB + APNB) method with other variants of our model in Fig.~\ref{fig:self}. In summary, our Full method shows the best semantic consistency and the least inconsistency artifacts while +AFNB and +APNB fail in some cases. The results also indicate AFNB and APNB is complementary to each other and the combination of them is beneficial to improve the performance.
	
	\vspace{1ex}\noindent \textbf{Selection of the fusing layers.} Fusing features from multi-levels is effective in many computer vision tasks. However, it still requires a lot of trials to find a good combination of the fusing layers. For semantic segmentation, the last several layers of the network contain plenty of features with semantic information, which is critical for better performance. Hence, we only combine the features from the shallow layers to the deep layers in a top-down manner. The responses are summed up if a certain layer receives more than two fusion invitations. The results are listed in Tab.~\ref{tab:fusing_ablation}. An obvious conclusion is that fusing only the features of $\mathrm{Stage4}$ and $\mathrm{Stage5}$ brings a considerable improvement while keep fusing more layers will only hurt the performance. 
	
	We conclude a possible intuitive reason: features from the early stages are generic to most tasks, while the last two are task-specific. Therefore, merging the features of the last several stages is more effective for a specific task. In the supplementary materials, we compare the feature visualizations of the outputs of all 5 stages of network trained on segmentation benchmark Cityscapes \cite{cityscapes} and of that trained on classification benchmark ImageNet \cite{Imagenet}, using the same backbone network ResNet-101 to demonstrate our guess.
	
\begin{figure*}[tb]
		\centering
		\includegraphics[width=0.95\textwidth]{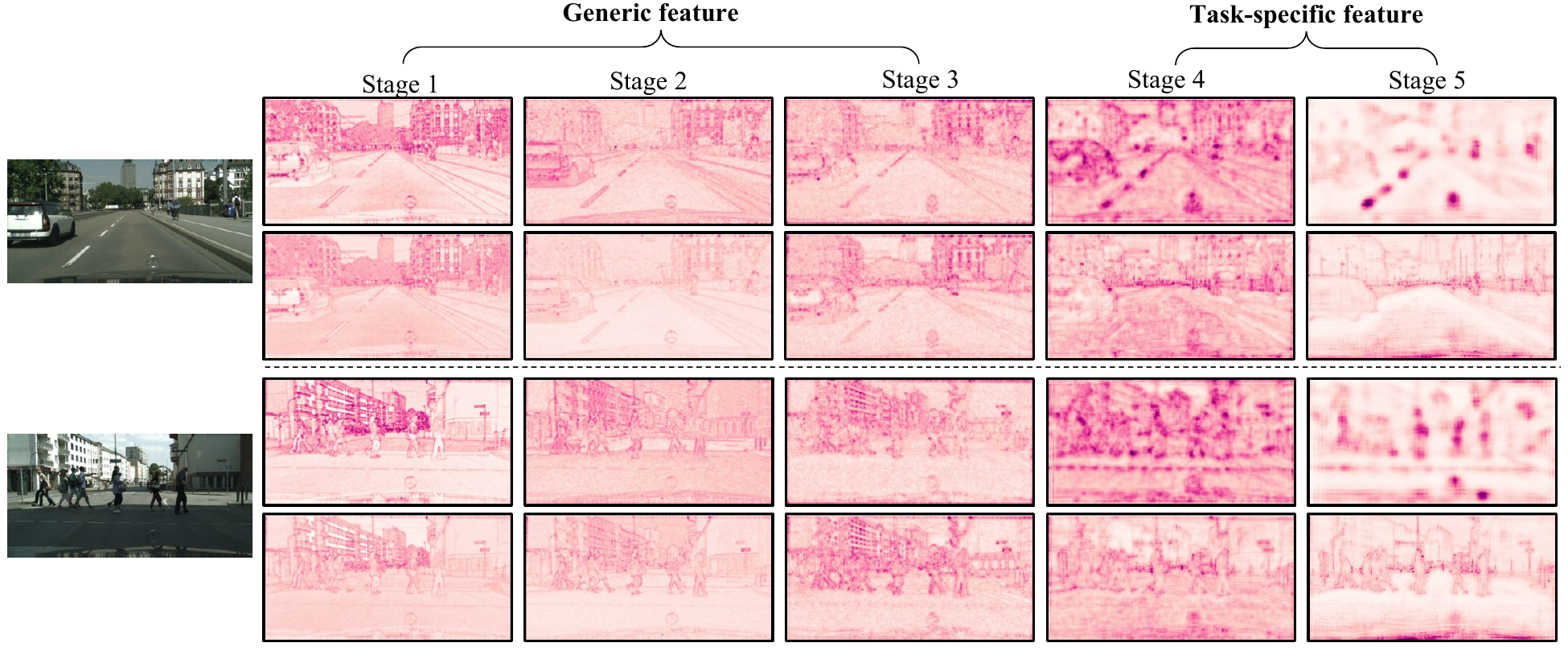}
		\caption{Feature visualization of different stages on ResNet-101. The \textbf{Upper} row represents visualizations from network trained on classification dataset ImageNet \cite{Imagenet}. The \textbf{Lower} row represents visualizations from network trained on scene segmentation dataset Cityscapes \cite{cityscapes}.}
		\label{fig:feature_vis}
	\end{figure*}

	As can be seen in Fig.~\ref{fig:feature_vis}, we compare the feature visualizations of the outputs of all 5 stages of network trained on segmentation benchmark Cityscapes \cite{cityscapes} (\textbf{{Lower}}) and of that trained on classification benchmark ImageNet \cite{Imagenet} (\textbf{{Upper}}), using the same backbone network ResNet-101. Comparing the two networks' feature visualizations of the same stage, the features are quite similar in the first three stages while differs hugely in the last two. This observation accord with our guess. Note our experiment results partially conforms to the conclusions in ExFuse \cite{ExFuse}, further validating the effectiveness of fusing only the last two stages.
	

	\begin{table}[b]
		\centering
		\begin{tabular}{l|c|c}
			\toprule[2pt]
			\hline
			Method          & Fusing layers  & mIoU (\%)  \\ \hline
			Baseline   & -- & 75.8  \\
			AFNB       &  4 \& 5      & 77.1   \\ 
			AFNB       &  3 \& 5, 4 \& 5      & 76.7   \\ 
			AFNB       &  2 \& 5, 3 \& 5, 4 \& 5      & 76.2   \\ \hline
			\bottomrule[2pt]
		\end{tabular}
		\caption{Ablation study on the validation set of Cityscapes in terms of the selection of layers to be fused.  ``4 \& 5'' means fusing the features of $\mathrm{Stage4}$ and $\mathrm{Stage5}$. Others likewise. \label{tab:fusing_ablation}}
	\end{table}
	
	\section{Discussions}
	We've conducted extensive experiments to summarize some experiences about semantic segmentation. As known, deep networks are fantastic tools but also need very careful tuning. The tuning process is extremely painful when it comes to semantic segmentation as there are many factors and the training process is very time and resources consuming. So we plan to share some experiences to the interested readers as references. But please do note these experiences are not as thorough as those in the main draft and it's very possible that some may not suit your cases. 
	\begin{itemize}
	   \item We found the types of graphic cards, pytorch versions, CUDA versions and NVIDIA driver versions may influence the performances. In our case, we found the same model trained on a workstation with 8 $\times $ Titan V is around 0.5 mIoU better than it trained on another machine with 8 $\times $ Titan Xp. 
	   \item The architecture of AFNB needs careful modifications to make AFNB work better. For example, we found adding a batch normalization layer after $W_o$ helps a lot when using AFNB alone while may not help on the Full model. We suspect there is a theoretical reason here and we're working to solve this issue. However, we found APNB is quite stable across a variaty of architectures. 
	   \item It seems that learning rate is important for segmentation models because we observe that the performance is boosted tremendously in the last 1/4 of training iterations. We also found that changing the initial learning rate has a great impact on the performance. We strongly suggest those, who are new to this task, refer to the learning rate configurations of the published methods that are mostly similar to your own method in architecture.
	   \item Using larger training iterations while keeping other configurations identical doesn't always improve the performance. In fact, changing training iterations is related to change the learning rate during training as we mainly adopt poly learning decay method in semantic segmentation. 
	\end{itemize}